\documentclass[runningheads]{llncs}

\usepackage[svgnames]{xcolor}

\usepackage{enumitem}
\usepackage{amsmath}
\usepackage{algpseudocode}

\usepackage[T1]{fontenc}

\usepackage[labelfont=bf,font=small]{caption}
\usepackage[labelfont=bf,font=small]{subcaption}
\usepackage{graphicx}
\usepackage{hyperref}

\usepackage[separate-uncertainty=true]{siunitx}
\usepackage{booktabs,makecell,multirow}
\renewrobustcmd{\boldmath}{}
\newrobustcmd{\B}{\fontseries{b}\selectfont}

\usepackage{tikz}

\makeatletter
\newcommand{\@chapapp}{\relax}%
\makeatother
\usepackage[title,toc,titletoc,header]{appendix}

\usepackage{comment}

\usepackage[linesnumbered,vlined,ruled,commentsnumbered]{algorithm2e}
\SetKwRepeat{Do}{do}{while}

\SetCommentSty{mycommfont}

\begin{document}
\title{FedGES: A Federated Learning Approach for Bayesian Network Structure Learning}
\titlerunning{FedGES: A Federated Learning Approach for BN Structure Learning}

\author{Pablo Torrijos\inst{1,2}\orcidID{0000-0002-8395-3848} \and
Jos\'e A. G\'amez\inst{1,2}\orcidID{0000-0003-1188-1117} \and
Jos\'e M. Puerta\inst{1,2}\orcidID{0000-0002-9164-5191}}

\authorrunning{P. Torrijos, JA. G\'amez and JM. Puerta}

%
\institute{Instituto de Investigaci\'on en Inform\'atica de Albacete (I3A). Universidad de Castilla-La Mancha. Albacete, 02071, Spain. \and
Departamento de Sistemas Inform\'aticos. Universidad de Castilla-La Mancha. Albacete, 02071, Spain.\\
\email{\{Pablo.Torrijos,Jose.Gamez,Jose.Puerta\}@uclm.es}}
\maketitle              
\begin{abstract}
Bayesian Network (BN) structure learning traditionally centralizes data, raising privacy concerns when data is distributed across multiple entities. This research introduces Federated GES (FedGES), a novel Federated Learning approach tailored for BN structure learning in decentralized settings using the Greedy Equivalence Search (GES) algorithm. FedGES uniquely addresses privacy and security challenges by exchanging only evolving network structures, not parameters or data. It realizes collaborative model development, using structural fusion to combine the limited models generated by each client in successive iterations. A controlled structural fusion is also proposed to enhance client consensus when adding any edge. Experimental results on various BNs from {\sf bnlearn}'s BN Repository validate the effectiveness of FedGES, particularly in high-dimensional (a large number of variables) and sparse data scenarios, offering a practical and privacy-preserving solution for real-world BN structure learning.

\keywords{Federated learning \and Bayesian Network structure learning \and Bayesian Network fusion/aggregation.}
\end{abstract}
%
%
%
%
%
\section{Introduction}\label{sec:introduction}
Bayesian Network (BN) structure learning \cite{deCamposLearningBN2011,chickering2002GES,gamezLearningBN2011} is a significant challenge in machine learning due to BNs' succinct depiction and interpretation of intricate probabilistic relationships \cite{Jensen2007BN}, being powerful tools for uncovering (in)dependence relationships in complex datasets. The recent interest in causal models \cite{Yao2021Causal} and the growing demand for explainable models \cite{BarredoArrieta2020XAI} have led to a widespread exploration of their applications and research.

These characteristics have allowed BNs to be adopted in areas such as agriculture \cite{Drury2017BNSurvey}, healthcare \cite{Kyrimi2021BNSurvey}, and renewable energy \cite{Borunda2016BNSurvey}. However, the computational complexity of learning BN structures becomes a significant challenge as the number of variables increases, as it is an NP-hard problem \cite{learningBNNP2004}.

In response, distributed learning efforts \cite{PGES2024,Laborda2023cGES,Scanagatta2019BNLearning} have divided the learning process among multiple nodes or clients. However, this approach raises privacy concerns, requiring a central node to access the entire dataset. In response, Federated Learning (FL) \cite{Li_review_FL_applications_2020,mcmahan17aFL,Zhang_survey_fl_KBS_2021} has emerged as a transformative paradigm, offering a collaborative framework for training machine learning models across privacy-sensitive environments. FL allows entities or clients to collaboratively learn a global model while maintaining data privacy locally, sharing only parameters, statistics, or model updates with a central server, which, if appropriately configured, do not compromise privacy.

Federated Learning's rapid development is evident in its applications in sectors facing data-sharing constraints, particularly in fields like healthcare \cite{Rieke2020FLSurvey,Silva2019FLSurvey} and fraud detection \cite{Yang2019FLSurvey}, where privacy concerns necessitate decentralized data processing. It also finds utility in scenarios involving client devices with limitations, such as low-powered Internet of Things (IoT) devices or mobile phones \cite{Leroy2019FLSurvey,mcmahan17aFL,Nguyen2021FLSurvey}. Additionally, it supports real-time systems where relying solely on server-sent models is impractical, as seen in applications like energy demand prediction \cite{Saputra2019FLSurvey}.



While FL has predominantly been applied to Neural Networks (NNs) \cite{Li_review_FL_applications_2020,Zhang_survey_fl_KBS_2021}, this study focuses on BNs, which offer interpretability lacking in the inherent black-box nature of NNs. Despite its potential, Federated Learning's application in Bayesian Network structure learning remains relatively unexplored. Existing literature in this domain typically employs horizontal data partitioning\footnote{Horizontal partitioning divides data instances across clients, where each client possesses complete records but for different samples or segments. In contrast, vertical partitioning splits data attributes across clients, with each retaining all instances but only for specific attributes or features.}. Methods range from adaptations of continuous optimization techniques \cite{ng22aNOTEARS} to approaches using regret-based learning \cite{mian22aRFCD,mian23aPERI}, and federated independence tests \cite{Wang2023FedC2SL}. These algorithms yield a global model comprising an unparameterized BN structure suitable for symbolic reasoning, such as relevance analysis. However, they often encounter challenges such as high execution times, generation of suboptimal BNs, or compromises in data privacy. In response, this paper proposes a novel Federated Learning BN structural learning paradigm based on the state-of-the-art Greedy Equivalence Search (GES) algorithm \cite{chickering2002GES} and Bayesian Network fusion techniques \cite{Puerta2021Fusion}.

\subsubsection{Contributions.} We present Federated GES (FedGES), a novel approach for Federated Bayesian Network structural learning in horizontally partitioned data settings. Our contributions include:
\begin{itemize}
    \item We present an iterative approach that combines (fuses) locally generated limited BNs from individual clients to construct a unified global structure. Our method ensures data security by exchanging only network structures or lists of (in)dependencies, thereby not exposing sensitive parameters (probabilities, statistics, etc.) between clients and the server.

    \item Our approach maintains the same theoretical properties as GES (identifying the optimal structure given sufficient and faithful data). FedGES includes mechanisms to ensure convergence and control the complexity of the fused structures, which are essential when these properties are challenged.
    
    \item We validate the efficacy of the proposed FedGES method through comprehensive experiments on various BNs from the {\sf bnlearn}'s Bayesian Network Repository \cite{bnlearn}. We focus on the final DAG obtained at the server, the global model, after the iterative process.

    \item The implementation of our algorithms is provided to ensure reproducibility and foster future research on this topic. 
\end{itemize}

\subsubsection{Organization of the paper.} Section \ref{sec:preliminaries} provides the necessary background and the related works. Section \ref{sec:fedGES} describes our proposed FedGES method in detail. Section \ref{sec:experiments} presents an experimental evaluation of our method using several benchmark BNs. Finally, Section \ref{sec:conclusions} concludes the paper and discusses future research directions.

%
%
\section{Preliminaries}\label{sec:preliminaries}

\subsection{Bayesian Network}\label{subsec:BN}
A Bayesian Network (BN) \cite{Jensen2007BN}, denoted as $\mathcal{B}=(\mathcal{G},\mathcal{P})$, is a probabilistic graphical model with two main components. On the one hand, a Directed Acyclic Graph (DAG), represented as $\mathcal{G}=(\mathcal{X},\mathcal{E})$, encapsulates the network structure, where $\mathcal{X}={X_1,\dots,X_n}$ denotes the problem domain variables and $\mathcal{E} = \{X_i \rightarrow X_j \mid X_i, X_j \in \mathcal{X} \land X_i \neq X_j\}$ encode the (in)dependency relationships between $\mathcal{X}$ through directed edges. On the other hand, a set of Conditional Probability Tables (CPTs), denoted by $\mathcal{P}$, factorizes the joint probability distribution $P(\mathcal{X})$ using the graphical structure $\mathcal{G}$ and the Markov's condition:
\begin{equation}
        P(\mathcal{X}) = P(X_1,\dots,X_n) = \prod_{i=1}^{n}P(X_i \; | \; pa_{\mathcal{G}}(X_i)),
\end{equation}
where $pa_{\mathcal{G}}(X_i)$ denotes the set of parents of $X_i$ in $\mathcal{G}$. In this paper, as usual in BN literature, we only consider discrete variables.

\subsection{Structure Learning of Bayesian Networks}\label{subsec:BNstructLearning}

Given a problem domain $\mathcal{X}=\{X_1,\dots,X_n\}$ and a dataset defined over it ${\cal{D}} = \{(x_1^i, \dots, x_n^i) \}_{i=1}^m$, the process of Bayesian Network structure learning \cite{deCamposLearningBN2011,chickering2002GES,gamezLearningBN2011} involves the derivation of a DAG ${\cal{G}} \in {\cal{G}}^n$, i.e. the space of DAGs defined over $n$ variables, such that ${\cal{G}}$ captures the relationships of (in)dependence among the variables in ${\cal{X}}$ supported by ${\cal{D}}$. Notably, BN structural learning poses an NP-hard problem \cite{learningBNNP2004}, necessitating the utilization of heuristic methods. Two primary approaches, namely constraint-based and score+search methods, are commonly employed in this context.

Constraint-based approaches, exemplified by the PC algorithm \cite{Spirtes2001PC}, aims to identify the skeleton of the underlying undirected graph by discerning dependencies among variables. This is accomplished through conditional independence tests, systematically removing edges inconsistent with the observed relations.

This work falls in the score+search approach, which seeks the optimal DAG $\mathcal{G}^* = \text{argmax}_{\mathcal{G}\in \mathcal{G}^n} f(\mathcal{G}:\mathcal{D})$, where $f(\mathcal{G}:\mathcal{D})$ is a scoring function that quantifies how well the DAG fits the given data $\mathcal{D}$. We employ the Bayesian Dirichlet equivalent uniform (BDeu) score \cite{heckerman1995BDeu}, a decomposable, score-equivalent metric that exhibits local and global consistency, characteristics for which it has been widely used over the years \cite{juanin2013GES,deCamposLearningBN2011,chickering2002GES,gamezLearningBN2011,PGES2024}. Several efficient local search-based algorithms, including the state-of-the-art algorithm, Greedy Equivalence Search (GES) \cite{chickering2002GES}, leverage these properties during the search process.

\subsubsection{Greedy Equivalence Search Algorithm}\label{subsec:ges}
The Greedy Equivalence Search (GES) algorithm \cite{chickering2002GES} stands out as a BN structural learning approach, showcasing high efficiency in practical applications. GES performs a greedy search within the space of equivalence classes of BN structures, consisting of two key phases: Forward Equivalence Search (FES) and Backward Equivalence Search (BES). During the FES stage, edges are incrementally added until a local maximum is reached. Conversely, edges are systematically deleted in the BES stage until a local optimum is attained. When coupled with a locally and globally consistent metric, the GES algorithm provides theoretical guarantees for identifying the optimal equivalence class given the data under sufficient and faithful data assumptions. However, practical considerations arise in the presence of certain substructures, necessitating modifications to reduce computational complexity while preserving the algorithm's theoretical properties \cite{juanin2013GES}. These adaptations enhance the algorithm's applicability and efficiency in real-world scenarios.


\subsection{Bayesian Network Structural Fusion} \label{subsec:BNFusion}
Bayesian Network Structural Fusion aims to construct a unified BN structure from a set of BNs that share the same variables. The objective is to synthesize input BNs by emphasizing their common independence relationships. Solving BN Fusion is a challenging task, being NP-hard, necessitating the utilization of heuristics for practical solutions. Heuristic approaches attempt to approximate the fusion of networks by relying on a common variable ordering $\sigma$ that may not be optimal \cite{penaConsensus2011}. A recent contribution introduces a practical and efficient greedy heuristic method for BN Fusion \cite{Puerta2021Fusion}. The proposed Greedy Heuristic Order (GHO) efficiently determines a suitable order $\sigma$ to guide the fusion process. The fusion process uses this order to obtain a minimal directed independence map\footnote{The minimal DAG $\mathcal{G}^{\sigma}$ that being compatible with $\sigma$ preserves as much as possible of the conditional independences in $\mathcal{G}$, although the number of arcs considerably increases.} $\mathcal{G}_i^{\sigma}$ following $\sigma$ for each input DAG $\mathcal{G}_i \in \{\mathcal{G}_1,\dots,\mathcal{G}_k\}$. Once the graphs are compatible with a common ordering, the consensus DAG is computed as the union of all the edges in $\{\mathcal{G}_1^{\sigma},\dots,\mathcal{G}_k^{\sigma}\}$. Authors in \cite{Puerta2021Fusion} show that the fusion obtained by using the heuristic ordering is almost identical to the optimal solution but requires, by far, less time.

\subsection{Federated Learning} \label{subsec:FL}
Since its introduction \cite{mcmahan17aFL}, Federated Learning (FL) \cite{Li_review_FL_applications_2020,Zhang_survey_fl_KBS_2021} has rapidly gained significance in scenarios involving privacy-sensitive and data-sharing constraints. Federated Learning can be defined as the process of learning a global model $\mathcal{M}$ through the collaboration of $k$ clients $\mathcal{C} = \{\mathcal{C}_1,\dots,\mathcal{C}_k\}$, each possessing its own dataset $\{\mathcal{D}_1,\dots,\mathcal{D}_k\}$ that is not shared with the other clients. If each database $\mathcal{D}_i$ contains the same set of $n$ variables $\mathcal{X} = \{X_1, \dots, X_n\}$, we would be talking about horizontal FL; otherwise, it would be vertical FL. 

Due to the high accuracy of deep learning approaches, most developments in Federated Learning involve deep neural networks. In this context, it is commonly assumed that the network structure is the same across all clients and the server; hence, only parameters (weights) are shared between the server and clients. The server aggregates the received parameters and dispatches the results to the clients, who then use the updated model alongside their own data to refine or update the model (weights). In the following, we will say that models are exchanged between the different nodes since, with the structure being common, the parameters differentiate the networks (models).

To summarize, the standard FL process is organized in a series of rounds. At each round: (1) The server $\mathcal{S}$ sends its model (initially random, empty, etc.) $\mathcal{M}$ to each client $\mathcal{C}_i$; (2) Each client $\mathcal{C}_i$ trains its local model $\mathcal{M}_i$ starting from $\mathcal{M}$ and using its respective dataset $\mathcal{D}_i$, and send it to the server; finally (3) the server aggregates all the received client's models $\{\mathcal{M}_1,\dots,\mathcal{M}_k\}$ to create a new global model $\mathcal{M}$, which becomes the starting point for the next iteration.

\subsection{Related Works} \label{subsec:relatedWorks}

There has been limited exploration of BNs in the FL domain. Featured FL approaches, like federated averaging \cite{mcmahan17aFL}, mainly focus on continuous optimization, typically applied to learning NNs. In this context, the NOTEARS-ADMM \cite{ng22aNOTEARS} algorithm emerges, adapting advances in continuous optimization to BN structure learning, specifically addressing horizontally partitioned data.

Furthermore, recent developments\footnote{It is important to highlight that, despite the terminology referring to structure learning of Causal Networks, in the three aforementioned contributions, we can use this term interchangeably with Bayesian Networks. This is attributed to their exploration of the space of Markov equivalence classes rather than the space of DAGs, highlighting their emphasis on equivalent causal structures.} have introduced methods based on both score+search algorithms, such as GES \cite{chickering2002GES}, and constraint-based algorithms, such as PC \cite{Spirtes2001PC}:
\begin{itemize}
    \item In the score+search category, notable contributions include the Regret-based Federated Causal Discovery (\textsc{RFcd}) \cite{mian22aRFCD} and its successor \textsc{Peri} \cite{mian23aPERI}. Both algorithms employ a regret-based search, where each client initially discovers its best-fitting local model using a score-based algorithm. The distinction lies in the search strategy, with \textsc{RFcd} utilizing a basic beam-search and \textsc{Peri} employing the GES algorithm. Following this, the server proposes networks, and clients return regret values relative to the proposed models. Utilizing these regrets, a global model is learned by minimizing the worst-case regret from all clients in a privacy-preserving manner.

    \item On the constraint-based side, the \textsc{FedC$^2$SL} algorithm \cite{Wang2023FedC2SL} stands out by introducing a federated framework for BN structure learning with a federated conditional independence $\chi^2$ test. This ensures an interaction with data that preserves privacy, enabling secure statistical evaluations of conditional independence between variables without sharing private data. By incorporating this federated framework, the \textsc{FedPC} algorithm emerges as an innovative approach to BN structure learning, extending the capabilities of the PC algorithm to the FL paradigm.
\end{itemize}

%
%
\section{FedGES} \label{sec:fedGES}

This section introduces Federated GES (FedGES), a novel approach to federated Bayesian Network structural learning. FedGES addresses the challenge of BN structure learning in privacy-sensitive environments by leveraging the Federated Learning paradigm. The algorithm follows a client-server framework, where the server orchestrates the fusion of client BNs into a global structure, ensuring privacy by exchanging only network structures (DAGs), not parameters. It is important to highlight that since no parameters or statistics are shared, a malicious agent cannot reconstruct or sample data from the Directed Acyclic Graph (DAG), even with access to it.

\begin{algorithm}[htbp]   
\setlength{\baselineskip}{1.3\baselineskip}
\caption{FedGES}\label{alg:fedGES}
\SetKwBlock{DoClients}{each client $i = 1,\dots,k$ do in parallel}{end}
\KwData{
$k$ clients; $\mathcal{D} = \{\mathcal{D}_1, \dots, \mathcal{D}_k\}$ datasets defined over $\mathcal{X} = \{X_1,\dots,X_n\}$ variables;
$maxIt$, the number of FL rounds;
$l$, the limit of edges that GES can add at each client in each FL round;
$fusionClients$ and $fusionServer$, the types of BN fusion the clients and the server execute, respectively.
}
\KwResult{
$\mathcal{G}=(\mathcal{V},\mathcal{E})$, the resulting DAG of the server; 
$\{\mathcal{G}_1,\dots,\mathcal{G}_k\} = \{(\mathcal{X},\mathcal{E}_1),\dots,(\mathcal{X},\mathcal{E}_k)\}$, the resulting client's DAGs.
}
\medskip

$\mathcal{G} \leftarrow (\mathcal{X}, \emptyset)$ \tcp*{Initialize server DAG}
\DoClients{
    $\mathcal{G}_i \leftarrow (\mathcal{X}, \emptyset)$ \tcp*{Initialize client-specific DAGs to the empty network}
}
\For{$(round=1,\dots,maxIt)$}{
    \DoClients{
        $\mathcal{G}_i' \leftarrow fusionClient(\mathcal{G}_i,\mathcal{G})$ \tcp*{Client-Side Fusion}
        $\mathcal{G}_i \leftarrow GES(init = \mathcal{G}_i', data=\mathcal{D}_i, limit=l)$ \tcp*{Obtain clients DAGs}
    }
    \If{convergenceCheck()}{
        break \tcp*{Convergence Verification}
    }
    $\mathcal{G} = fusionServer(\{\mathcal{G}_1,\dots,\mathcal{G}_k\})$ \tcp*{Server-Side Fusion}
}
\end{algorithm}

The FedGES scheme is outlined in Algorithm \ref{alg:fedGES}. Let $\{\mathcal{D}_1, \dots, \mathcal{D}_k\}$ represent $k$ unique datasets, one for each of the $k$ clients, defined on the same set of variables $\mathcal{X} = \{X_1,\dots,X_n\}$. The server starts with an empty DAG, $\mathcal{M}$ = $\mathcal{G}$ = $(\mathcal{X},\emptyset)$, and the models $\{\mathcal{M}_1, \dots, \mathcal{M}_k\} = \{\mathcal{G}_1, \dots, \mathcal{G}_k\}$ represent the specific DAG of each client, initially also empty. 

The process begins with each client fusing the received global DAG $\mathcal{G}$ with its DAG $\mathcal{G}_i$, obtaining $\mathcal{G}_i'$. Any type of fusion can be used for this, or even the incoming DAG can overwrite the local one ($\mathcal{G}_i' = \mathcal{G}$). 
At each round of the FL cycle, each client obtains $\mathcal{G}_i$ running a BN learning algorithm which takes $\mathcal{G}_i'$ as initial state and is \textit{constrained to add a limited number of edges, $l$}, thus facilitating a more gradual learning process that avoids excessively complex fusions in the server. In this paper, the GES algorithm with the BDeu score (Section \ref{subsec:BNstructLearning}) is employed as the locally used structural learning algorithm. Then, the locally learned models $\{\mathcal{G}_1,\dots,\mathcal{G}_k\}$ are then sent to the server, which fuses them into a single model $\mathcal{G}$ by using the fusion method proposed in \cite{Puerta2021Fusion}. Therefore, in this approach, uniform fusion is used, and after the fusion, the same (global) model is sent to all the clients.
This iterative process continues until the maximum number of rounds ($maxIt$) or when none of the new local networks $\mathcal{G}_i$ differs from the previous iteration. The latter criterion can be used when the BN learning algorithm is deterministic, as with GES.

The proposed approach offers several advantages. Firstly, the framework is highly customizable, allowing adjustments to the base structural learning algorithm or its score function and modifications to the fusion process on both the client and server sides. Secondly, this approach maintains the same good theoretical properties as the GES algorithm when using a globally and locally consistent scoring criterion as the BDeu score \cite{chickering2002GES}. It asymptotically converges to the gold-standard BN $(\mathcal{G}^*,\theta^*)$ under the same assumptions. This convergence is demonstrated by the fact that given sufficient data faithful to the probability distribution $\theta^*$ encoded by the optimal BN, the GES algorithm for each client $i$ returns a learned model $\mathcal{G}_i$ equal to $\mathcal{G}^*$. Consequently, the subsequent server fusion, $\mathcal{G}$, also converges to $\mathcal{G}^*$. This holds when using a fusion method that, given a list of equal DAGs $\{\mathcal{G},\dots,\mathcal{G}\}$, always results in $\mathcal{G}$ (as is the case with all the methods used in this paper).

The convergence of each GES run on clients is guaranteed by employing a consistent scoring criterion, ensuring that each $\mathcal{G}_i$ score is greater than or equal to the initial $\mathcal{G}_i'$ score. However, the overall convergence of FedGES is not assured because the resulting $\mathcal{G}_i'$ from fusing the last $\mathcal{G}_i$ with the global $\mathcal{G}$ may obtain a lower score than $\mathcal{G}_i$. While this can potentially enhance the quality of generated Bayesian Networks (BNs), as incorporating information from other networks may lead to a score decrease, it can also result in cyclical behavior. Some clients might repeat the same DAGs every certain number of iterations. This fact can be addressed by modifying the convergence function to check whether any of the $\mathcal{G}_i$ generated in the current iteration has not been created earlier in previous iterations rather than checking only for changes with respect to the previous iteration.

%
%
\section{Experimental Evaluation} \label{sec:experiments}
This section presents the practical evaluation of our proposal compared to alternative hypotheses. We discuss the domains and algorithms involved, detail the experimental approach, and present the obtained results.

\subsection{Algorithms} \label{subsec:algorithms}
The algorithms evaluated in this study include\footnote{We ran algorithms for which the source code is publicly available. NOTEARS-ADMM was not included in our tests, as previous works \cite{mian22aRFCD,mian23aPERI} have demonstrated its inferior performance compared to the other methods we used to evaluate FedGES.}:

\begin{itemize}
    \item The \textsc{RFcd} algorithm \cite{mian22aRFCD}.
    \item The \textsc{FedC2SL} algorithm \cite{Wang2023FedC2SL} instantiated in \textsc{FedPC}. We also include in the comparison their baseline algorithm PC-Voting \cite{Wang2023FedC2SL}.
    \item An enhanced version of the GES algorithm \cite{chickering2002GES} described in \cite{juanin2013GES}. Although not initially designed as a federated algorithm, we learn a network for each client and fuse them using the procedure outlined in \cite{Puerta2021Fusion}, obtaining a baseline for the global ($\mathcal{G}$) model. Notice that in this method, GES is run at each client until convergence; that is, no limit on the number of edges is set. This baseline method can be viewed as a one-shot FL algorithm because only a single round of communication is needed.
    \item The proposed FedGES algorithm (see Section \ref{sec:fedGES}), configured with a limit of $l=10$ edges added by GES at each round. Three fusion strategies are tested on the server side: The canonical BN fusion described in \cite{Puerta2021Fusion}, denoted as \textsc{Union}, which adds the edges of all the BNs once converted to the same $\sigma$ ancestral order; and \textsc{C25} and \textsc{C50}, which only add those edges that appear in at least 25\% or 50\% of input DAGs transformed to $\sigma$, respectively. This ensures a minimum consensus among different clients when adding each arc while also limiting the complexity (number of arcs) of the resulting fused model.
\end{itemize}

\subsection{Methodology} \label{subsec:methodology}
To assess the validity of FedGES, we have selected 14 BNs from the {\sf bnlearn}'s Bayesian Network Repository\footnote{\href{https://www.bnlearn.com/bnrepository/}{https://www.bnlearn.com/bnrepository/}} \cite{bnlearn}, generating 10 samples (datasets) with 5000 instances for each one. The focus is on BNs of Medium, Large, and Very Large sizes, described in Table \ref{table-BNs}. Small BNs are excluded intentionally.
The selected BNs constitute a commonly used benchmark in the BN learning literature.

In the experimental evaluation, each sample of 5000 instances is distributed among the varying number of clients used in each algorithm (5, 10, or 20). Consequently, each client is allocated 1000, 500, or 250 instances, respectively. This design allows us to study scenarios with limited data, where Federated Learning becomes more valuable in real-world applications.

\begin{table}[htb]  
\caption{Bayesian Networks used in the experimental evaluation.}
\label{table-BNs}
\resizebox{\textwidth}{!} {%
\begin{tabular*}{1.2\textwidth}{@{\extracolsep{\fill}}lS[table-format=4.0]S[table-format=4.0]S[table-format=7.0]S[table-format=1.0]S[table-format=4.0]}
\toprule
\multicolumn{1}{c}{\multirow{2}{*}{\textsc{\bfseries Network}}} &\multicolumn{5}{c}{\textsc{\bfseries Features}} \\
\cmidrule(){2-6}
& \multicolumn{1}{r}{\textsc{Nodes}} & \multicolumn{1}{r}{\textsc{Edges}} & \multicolumn{1}{r}{\textsc{Parameters}} & \multicolumn{1}{r}{\textsc{Max parents}} & \multicolumn{1}{r}{\textsc{Empty SMHD}}\\
\midrule
\textsc{Child}           & 20	& 25 & 230 & 4 & 30 \\
\textsc{Insurance}       & 27	& 52 & 1008 & 3 & 70 \\
\textsc{Water}           & 32	& 66 & 10083 & 5 & 123 \\
\textsc{Mildew}          & 35	& 46 & 540150 & 3 & 80 \\
\textsc{Alarm}           & 37	& 46 & 509 & 4 & 65 \\
\textsc{Barley}          & 48	& 84 & 114005 & 4 & 126 \\
\midrule
\textsc{Hailfinder}      & 56	& 66 & 2656 & 4 & 99 \\
\textsc{Hepar2}          & 70	& 123 & 1453 & 6 & 158 \\
\textsc{Win95pts}        & 76	& 112 & 574 & 7 & 225 \\
\midrule
\textsc{Pathfinder}      & 109 & 195 & 72079 & 5 & 208 \\
\textsc{Andes}           & 223	& 338 & 1157 & 6 & 626 \\
\textsc{Diabetes}        & 413 & 602 & 429409 & 2 & 819 \\
\textsc{Pigs}            & 441 & 592 & 5618 & 2 & 806 \\
\textsc{Link}            & 724 & 1125 & 14211 & 3 & 1739 \\
\bottomrule
\end{tabular*}
}
\end{table}

To evaluate the networks obtained by using the tested methods, this article only compares the structure of the discovered networks because using learning scores as BDeu only makes sense on the client's side, as the server has no access to any data.
In particular, the Structural Moralized Hamming Distance (SMHD) \cite{Kim2019SHDMoral,Laborda2023cGES,PGES2024} is used to assess the similarity of the learned BNs to the original one from which the data were sampled. This metric is akin to the literature's widely used Structural Hamming Distance (SHD) \cite{SHD2009}, but it compares the moralized graphs to avoid considering different equivalent (in)dependencies. As the SMHD compares the structure of the BNs, we can use this metric to assess the quality of networks obtained by the clients and also by the server.

\subsection{Reproducibility} \label{subsec:reproducibility}
To ensure reproducibility, we implemented the GES algorithm and the federated framework from scratch using Java (OpenJDK 20) and the Tetrad 7.1.2-2\footnote{\href{https://github.com/cmu-phil/tetrad/releases/tag/v7.1.2-2}{https://github.com/cmu-phil/tetrad/releases/tag/v7.1.2-2}} causal reasoning library. For comparisons with \textsc{FedC$^2$SL} and \textsc{RFcd}, we use the Python implementation of both algorithms as provided by the authors of \textsc{FedC$^2$SL}, available on GitHub\footnote{\href{https://github.com/wangzhaoyu07/FedC2SL}{https://github.com/wangzhaoyu07/FedC2SL}}, and executed them with Python 3.10.8. All experiments were conducted on machines equipped with Intel Xeon E5-2650 8-Core Processors and 64 GB of RAM per execution.

Furthermore, for transparency and accessibility, we have created a shared repository on OpenML\footnote{\href{https://www.openml.org/search?type=data\&uploader\_id=\%3D\_33148\&tags.tag=bnlearn}{https://www.openml.org/search?type=data\&uploader\_id=\%3D\_33148\&tags.tag=bnlearn}} containing the 10 datasets sampled for each of the 14 BNs, with references to their original papers included in their descriptions. These databases, along with all source code, are also available on GitHub\footnote{\href{https://github.com/ptorrijos99/BayesFL}{https://github.com/ptorrijos99/BayesFL}}.

\subsection{Results} \label{subsec:results}
\begin{figure}[t]   
    \centering
    \includegraphics[width=1\linewidth]{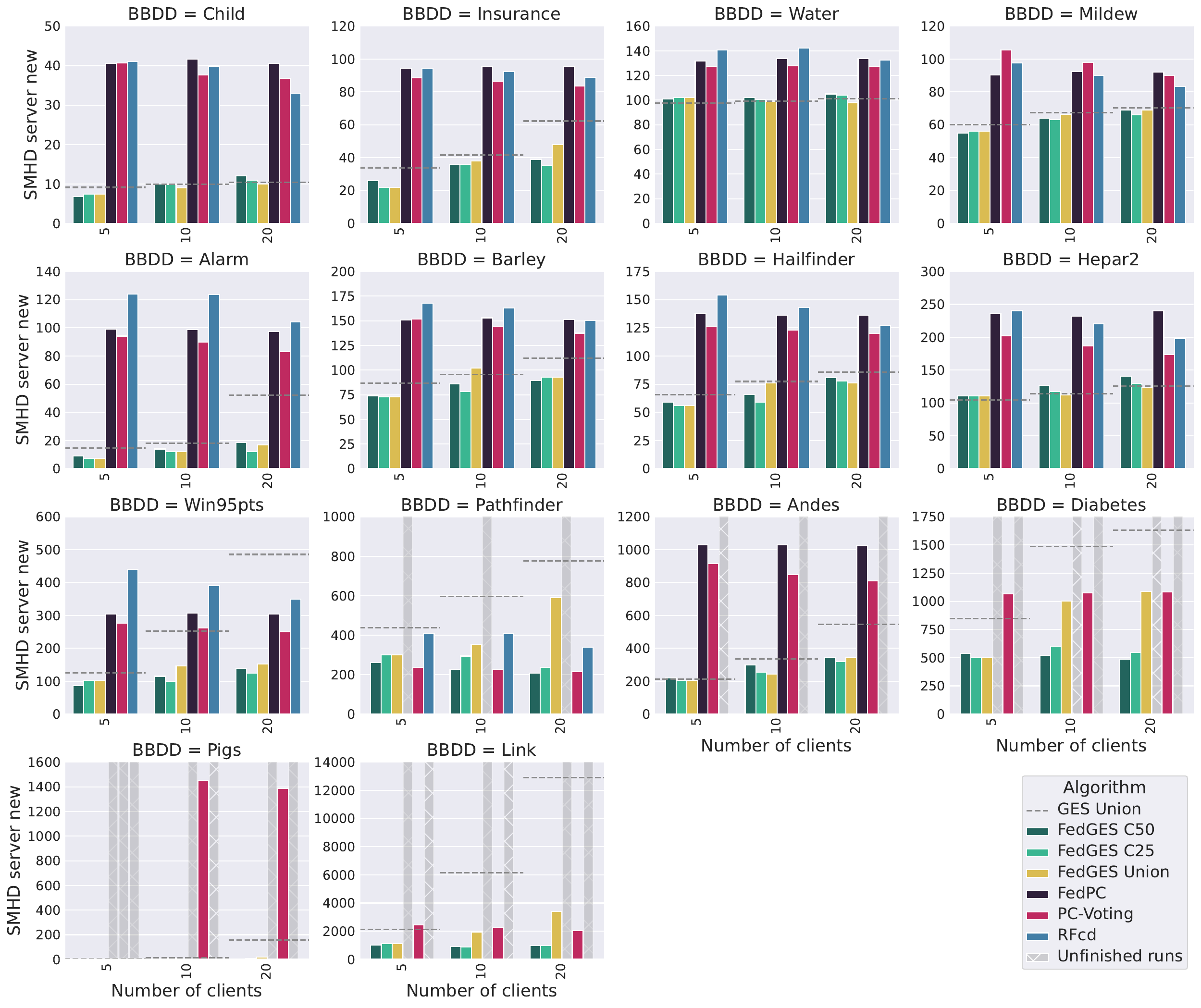}
    \caption{Mean SMHD of the final global BNs $\mathcal{G}$. Comparison with baselines. ``GES Union'' lines correspond to running the GES algorithm with no iteration limit on all clients and fusing the resulting networks with a \textsc{Union}. The ``Unfinished run'' bars correspond to algorithms that cannot finish the execution in a reasonable amount of time.}
    \label{fig:SMHDserver-baselines}
\end{figure}

In this section, we assess the performance of FedGES by evaluating the global DAG ($\mathcal{G}$) obtained after the federated process. 
The SMHD scores of the global BNs $\mathcal{G}$ generated by the server for each algorithm and configuration (\#clients, fusion) are presented in Figure \ref{fig:SMHDserver-baselines} \footnote{With 5 clients, \textsc{C25} and \textsc{Union} fusion are equivalent ($\lfloor5 \cdot 0.25\rfloor = \lfloor1.25\rfloor = 1$), involving the addition of all edges present in the DAGs.}. Additionally, Table \ref{tab:result-edges-server} provides the count of added edges in these networks. 

\begin{table}[htbp]
\caption{Mean number of edges of the final networks generated by the server.}
\label{tab:result-edges-server}
\resizebox{\linewidth}{!} {%
\begin{tabular*}{1.6\textwidth}{@{\extracolsep{\fill}}l*{11}{S[table-format=3.1]}@{}}
\toprule
\multirowcell{2}{\textsc{\bfseries Network}}  & \multicolumn{3}{c}{\textsc{\bfseries 5 Clients}} & \multicolumn{4}{c}{\textsc{\bfseries 10 Clients}} & \multicolumn{4}{c}{\textsc{\bfseries 20 Clients}} \\
\cmidrule(){2-4}\cmidrule(){5-8}\cmidrule(){9-12}
 & \textsc{C50} & \textsc{C25/Union} & \textsc{GES} & \textsc{C50} & \textsc{C25} & \textsc{Union} & \textsc{GES} & \textsc{C50} & \textsc{C25} & \textsc{Union} & \textsc{GES} \\
\midrule
\textsc{Child}      &   21.6 &    21.8 &    21.6 &   20.0 &   20.9 &   23.0 &    22.4 &   17.0 &   18.0 &    23.0 &    25.9 \\
\textsc{Insurance}  &   37.5 &    45.0 &    48.5 &   32.0 &   43.0 &   48.0 &    57.4 &   29.0 &   32.0 &    54.0 &    68.2 \\
\textsc{Water}      &   25.0 &    31.0 &    35.9 &   20.0 &   23.0 &   31.0 &    40.8 &   18.0 &   19.0 &    31.0 &    40.2 \\
\textsc{Mildew}     &   28.0 &    29.0 &    32.0 &   23.0 &   27.0 &   29.0 &    31.9 &   17.0 &   21.0 &    26.0 &    26.8 \\
\textsc{Alarm}      &   44.0 &    48.5 &    55.0 &   42.0 &   46.0 &   53.0 &    61.4 &   38.5 &   44.0 &    53.5 &    90.8 \\
\textsc{Barley}     &   49.0 &    52.0 &    62.2 &   40.0 &   47.0 &   52.0 &    58.1 &   35.5 &   47.0 &    47.0 &    62.6 \\
\cmidrule(){1-1}\cmidrule(){2-4}\cmidrule(){5-8}\cmidrule(){9-12}
\textsc{Hailfinder} &   62.0 &    68.0 &    81.9 &   46.0 &   59.0 &   73.0 &    84.1 &   35.0 &   44.0 &    58.7 &    74.4 \\
\textsc{Hepar2}     &   43.5 &    51.0 &    54.6 &   30.0 &   39.5 &   50.5 &    52.4 &   16.0 &   27.0 &    51.3 &    57.2 \\
\textsc{Win95pts}   &  108.0 &   136.0 &   176.4 &   87.0 &  109.0 &  150.0 &   262.5 &   70.0 &   92.0 &   150.0 &   392.4 \\
\cmidrule(){1-1}\cmidrule(){2-4}\cmidrule(){5-8}\cmidrule(){9-12}
\textsc{Pathfinder} &  135.5 &   170.0 &   298.2 &  103.0 &  155.0 &  200.0 &   414.4 &   75.5 &  116.2 &   305.0 &   534.1 \\
\textsc{Andes}      &  255.5 &   273.0 &   267.8 &  216.0 &  242.0 &  295.0 &   339.0 &  193.0 &  220.3 &   330.0 &   437.5 \\
\textsc{Diabetes}   &  561.0 &   620.0 &   980.6 &  499.0 &  577.5 &  646.2 &  1451.9 &  462.2 &  533.0 &   846.0 &  1463.7 \\
\textsc{Pigs}       &  592.0 &   601.0 &   600.5 &  593.0 &  600.0 &  600.0 &   607.5 &  592.0 &  594.0 &   622.5 &   764.4 \\
\textsc{Link}       &  856.0 &  1012.0 &  2108.7 &  778.2 &  959.0 &  655.5 &  3947.1 &  642.0 &  819.8 &  2211.0 &  6386.4 \\
\midrule
\end{tabular*}
}

\resizebox{\linewidth}{!} {%
\begin{tabular*}{1.6\textwidth}{@{\extracolsep{\fill}}l*{9}{S[table-format=3.1]}@{}}
\toprule
\multirowcell{2}{\textsc{\bfseries Network}}  & \multicolumn{3}{c}{\textsc{\bfseries 5 Clients}} & \multicolumn{3}{c}{\textsc{\bfseries 10 Clients}} & \multicolumn{3}{c}{\textsc{\bfseries 20 Clients}} \\
\cmidrule(){2-4}\cmidrule(){5-7}\cmidrule(){8-10}
 & \textsc{FedPC} & \textsc{PC-Voting} & \textsc{RFcd} & \textsc{FedPC} & \textsc{PC-Voting} & \textsc{RFcd} &\textsc{FedPC} & \textsc{PC-Voting} & \textsc{RFcd} \\
\midrule
\textsc{Child}      &   27.2 &   24.7 &   28.8 &   27.0 &   21.3 &   25.5 &   26.0 &   21.4 &   16.2 \\
\textsc{Insurance}  &   46.8 &   38.7 &   47.5 &   48.2 &   36.4 &   45.5 &   48.0 &   27.1 &   39.8 \\
\textsc{Water}      &   31.6 &   19.0 &   55.8 &   32.2 &   18.1 &   47.8 &   34.2 &   17.0 &   35.8 \\
\textsc{Mildew}     &   20.2 &   42.7 &   31.0 &   23.4 &   33.3 &   22.5 &   22.6 &   23.1 &   12.8 \\
\textsc{Alarm}      &   54.2 &   47.1 &   86.0 &   53.2 &   42.0 &   85.5 &   52.0 &   32.1 &   62.5 \\
\textsc{Barley}     &   56.0 &   51.9 &   74.2 &   58.8 &   41.0 &   70.7 &   57.2 &   30.4 &   48.3 \\
\cmidrule(){1-1}\cmidrule(){2-4}\cmidrule(){5-7}\cmidrule(){8-10}
\textsc{Hailfinder} &   56.0 &   42.9 &   71.5 &   54.8 &   38.1 &   58.0 &   55.4 &   34.4 &   37.0 \\
\textsc{Hepar2}     &   95.4 &   53.9 &  103.2 &   92.0 &   35.9 &   73.5 &  101.4 &   21.3 &   56.0 \\
\textsc{Win95pts}   &  105.2 &   69.7 &  264.5 &  108.0 &   48.1 &  209.5 &  104.8 &   33.6 &  156.0 \\
\cmidrule(){1-1}\cmidrule(){2-4}\cmidrule(){5-7}\cmidrule(){8-10}
\textsc{Pathfinder} &    {\ \ -} &   41.4 &  237.0 &    {\ \ -} &   25.8 &  227.0 &    {\ \ -} &   14.3 &  149.0 \\
\textsc{Andes}      &  430.8 &  310.9 &    {\ \ -} &  429.0 &  240.7 &    {\ \ -} &  423.8 &  194.4 &    {\ \ -} \\
\textsc{Diabetes}   &    {\ \ -} &  260.0 &    {\ \ -} &    {\ \ -} &  268.0 &    {\ \ -} &    {\ \ -} &  276.5 &    {\ \ -} \\
\textsc{Pigs}       &    {\ \ -} &    {\ \ -} &    {\ \ -} &    {\ \ -} &  664.0 &    {\ \ -} &    {\ \ -} &  592.5 &    {\ \ -} \\
\textsc{Link}       &    {\ \ -} &  704.5 &    {\ \ -} &    {\ \ -} &  504.7 &    {\ \ -} &    {\ \ -} &  304.3 &    {\ \ -} \\
\bottomrule
\end{tabular*}
}
\end{table}

The obtained results lead to several key conclusions:
\begin{itemize}
    \item FedGES consistently outperforms other algorithms in all scenarios, except for the \textsc{Pathfinder} network, where, in specific instances, the PC-Voting and \textsc{RFcd} algorithms achieve slightly better results. This divergence can be attributed to the unique characteristics of the \textsc{Pathfinder} network, which has a semi-naive Bayes structure\footnote{\href{https://www.bnlearn.com/bnrepository/discrete-verylarge.html\#pathfinder}{https://www.bnlearn.com/bnrepository/discrete-verylarge.html\#pathfinder}}, where improvements in the BDeu score do not necessarily translate to better SMHD \cite{PGES2024}. As a result, neither algorithm enhances the SMHD of an empty network (208). When analyzing the number of added edges, it is evident that all algorithms achieve a better SMHD when adding fewer edges. Furthermore, this count decreases with an increasing number of clients. This dynamic favors PC-Voting, returning a nearly empty network. In all other networks, the performance of \textsc{FedPC}, PC-Voting, and \textsc{RFcd} is very poor, with PC-Voting surprisingly outperforming \textsc{FedPC} in virtually all BNs. Also noteworthy is the case of \textsc{Pigs}, wherein the two cases where PC-Voting terminates (10 and 20 clients), it obtains very poor results in SMHD. At the same time, the three FedGES configurations practically attain the optimal network.
    
    \item The two competing algorithms specifically designed for FL, \textsc{FedPC} and \textsc{RFcd}, fail to run on large BNs. This fact highlights the relevance of FedGES, especially in scenarios involving large networks.

    \item The application of Federated Learning with FedGES improves the results of merging BNs generated by non-federated GES. This improvement becomes more pronounced as the size of the BN increases and when more clients (each one with less data) are involved. It is logical since clients with smaller datasets introduce more variability in the networks they generate. As a result, the fusion of all these networks may contain numerous unnecessary edges, resulting in very poor SMHDs. This effect is more significant in larger and more challenging networks such as \textsc{Pathfinder}, \textsc{Diabetes}, or \textsc{Link} with more possible edges to add.

    \item Comparing the three fusions carried out in FedGES (C50, C25, and \textsc{Union}), it is evident that the more the BNs generated by the clients diverge among them, whether through an increase in the number of clients or through the creation of larger BNs (as seen in the previous case), the more advisable it becomes to utilize C50 or C25. This is logical since, in small BNs and with a large amount of data, the BNs generated by each client will not differ enough for a C50 or C25 strategy to be noticeable. Furthermore, with 5 clients, C25 is equivalent to \textsc{Union}, and C50 actually adds each edge if it appears in 2 of the 5 networks, which is quite likely. However, in the most complex BNs, we can clearly see how the use of C50 or C25 produces very good results when the number of clients increases. Both \textsc{Union} and GES increase the number of edges added, and so the SMHD value, which in some cases is even worse than that of PC-Voting or \textsc{RFcd}.
\end{itemize}

\begin{table}[htbp]
\caption{Total execution time (seconds) normalized by the number of clients.}
\label{tab:result-time}
\resizebox{\linewidth}{!} {%
\begin{tabular*}{2.1\textwidth}{@{\extracolsep{\fill}}lS[table-format=4.1]S[table-format=4.1]S[table-format=3.1]S[table-format=3.1]S[table-format=4.1]S[table-format=5.1]
S[table-format=4.1]S[table-format=4.1]S[table-format=4.1]S[table-format=3.1]S[table-format=3.1]S[table-format=4.1]S[table-format=4.1]
S[table-format=3.1]S[table-format=4.1]S[table-format=3.1]S[table-format=3.1]S[table-format=3.1]S[table-format=4.1]S[table-format=4.1]}
\toprule
\multirowcell{2}{\textsc{\bfseries Network}}  & \multicolumn{6}{c}{\textsc{\bfseries 5 Clients}} & \multicolumn{7}{c}{\textsc{\bfseries 10 Clients}} & \multicolumn{7}{c}{\textsc{\bfseries 20 Clients}} \\
\cmidrule(){2-7}\cmidrule(){8-14}\cmidrule(){15-21}
 & \textsc{C50} & \textsc{C25/Un} & \textsc{GES}  & \textsc{FedPC} & \textsc{PC-Vot} & \textsc{RFcd} & \textsc{C50} & \textsc{C25} & \textsc{Union} & \textsc{GES}  & \textsc{FedPC} & \textsc{PC-Vot} & \textsc{RFcd} & \textsc{C50} & \textsc{C25} & \textsc{Union} & \textsc{GES}  & \textsc{FedPC} & \textsc{PC-Vot} & \textsc{RFcd} \\
\midrule
\textsc{Child}      &     \B 0.2&     \B 0.2&    \B 0.2&    7.2 &     1.7 &     30.5 &    \B 0.1&    \B 0.1&    \B 0.1&   \B 0.1&    4.5 &     1.2 &    18.3 &    0.1&     0.1&    0.1&   \B 0.0&    3.3 &      0.9&    12.5 \\
\textsc{Insurance}  &     \B 0.2&      0.3&    \B 0.2&   11.2 &     3.2 &    106.3 &      0.2&      0.4&     0.7&   \B 0.1&    6.9 &     2.3 &    56.5 &     0.3&      0.3&   11.0 &   \B 0.1&    4.7 &     1.5&    34.8 \\
\textsc{Water}      &      0.3&      0.3&    \B 0.2&     2.8&      0.9&    140.8 &    \B 0.1&      0.4&      0.2&   \B 0.1&    1.7 &     0.7&    78.5 &   \B 0.1&    \B 0.1&     0.3&   \B 0.1&    1.1 &      0.5&    43.4 \\
\textsc{Mildew}     &     1.0 &     1.0 &    \B 0.4&  172.1 &    11.3 &    270.9 &     1.4 &     1.2 &    19.3 &    \B 0.2&  130.8 &     5.4 &   176.5 &    1.1 &      0.9&    7.6 &   \B 0.1&  102.3 &     3.0 &    96.9 \\
\textsc{Alarm}      &      0.4&      0.4&    \B 0.3&    8.4 &     3.6 &    328.1 &      0.5&      0.3&      0.3&    \B 0.2&    5.0 &     3.0 &   165.4 &     0.2&      0.4&    1.4 &   \B 0.1&    3.4 &     2.4 &    88.6 \\
\textsc{Barley}     &     1.3 &     1.5&    \B 0.5&   67.5 &     7.8 &    820.8 &     3.1 &    16.4 &    16.3 &    \B 0.3&   42.9 &     5.3 &   468.9 &   12.3 &    22.4 &   28.7 &    \B 0.2&   31.1 &     3.6 &   276.2 \\
\cmidrule(){1-1}\cmidrule(){2-7}\cmidrule(){8-14}\cmidrule(){15-21}
\textsc{Hailfinder} &     0.8 &     1.4 &    \B 0.4&   17.8 &     5.6 &    980.8 &     0.6 &     1.7 &     3.1 &    \B 0.3&   10.6 &     4.4 &   480.2 &    0.7&     0.4&     2.8&    \B 0.2&    7.0 &     3.5 &   216.3 \\
\textsc{Hepar2}     &     0.6 &     1.4 &    \B 0.5&   34.0 &     6.1 &   2727.9 &     \B 0.4&     1.4 &     3.7 &    \B 0.3&   21.6 &     4.2 &  4609.1 &    \B 0.2&      0.3&    2.9 &    \B 0.2&   15.1 &     3.2 &  1441.3 \\
\textsc{Win95pts}   &     2.2 &     1.6 &    \B 0.9&   25.0 &    12.2 &   6248.5 &     1.1 &    \B 1.5&     3.1 &   \B 0.7&   14.9 &    10.1 &  2185.9 &    1.6 &     2.0&  251.2 &    \B 0.5&    9.7 &     8.0 &   884.3 \\
\cmidrule(){1-1}\cmidrule(){2-7}\cmidrule(){8-14}\cmidrule(){15-21}
\textsc{Pathfinder} &   155.8 &   367.5 &    \B 2.8&    {\ \ -} &  2564.3 &  13972.9 &    15.3 &    13.3 &   114.2 &   \B 2.0&    {\ \ -} &   308.6 &  6625.7 &    7.6 &    24.0 &   37.2 &   \B 1.5&    {\ \ -} &    81.5 &  2941.2 \\
\textsc{Andes}      &    15.7 &    \B 12.6&   16.3 &  215.4 &    75.0 &      {\ \ -} &    14.7 &    15.1 &    13.9 &  \B 13.8&  132.0 &    57.3 &     {\ \ -} &   17.6 &    22.3 &   73.2 &  \B 10.9&   92.4 &    47.7 &     {\ \ -} \\
\textsc{Diabetes}   &   786.5 &   746.2 &  \B 154.3&    {\ \ -} &  6119.2 &      {\ \ -} &   601.3 &   665.5 &  1313.8 & \B 148.2&    {\ \ -} &  2694.3 &     {\ \ -} &  677.4 &   518.5 &  511.0 & \B 120.2&    {\ \ -} &  1614.9 &     {\ \ -} \\
\textsc{Pigs}       &   106.8 &    \B 70.1&  150.2 &    {\ \ -} &     {\ \ -} &      {\ \ -} &   129.4 &   \B 68.9&    96.8 &  147.9 &    {\ \ -} &  3320.6 &     {\ \ -} &  131.5 &   \B 90.7&   96.9 &  143.6 &    {\ \ -} &   699.3 &     {\ \ -} \\
\textsc{Link}       &  3186.2 &  1811.7 &  \B 656.0&    {\ \ -} &  2925.1 &      {\ \ -} &  1343.6 &  1226.2 &  \B 543.9&  608.8 &    {\ \ -} &  1845.9 &     {\ \ -} &  608.5 &  1235.1 &  513.6 & \B 504.2&    {\ \ -} &  1070.1 &     {\ \ -} \\
\bottomrule
\end{tabular*}
}
\end{table}

Finally, Table \ref{tab:result-time} presents the total execution times (sum of all clients and the server) for various algorithms, normalized by the number of clients for clarity. It is evident that \textsc{FedPC}, PC-Voting, and \textsc{RFcd} exhibit computational complexities several orders of magnitude higher than the FedGES configurations and GES, rendering them impractical for larger Bayesian Networks.

When comparing FedGES with different fusion strategies, we observe that with fewer clients (5), resulting in fewer and more consistent BNs due to increased data per client, the \textsc{Union} fusion (equivalent to C25) generally outperforms C50 in larger BNs. As the number of clients increases (10, 20), more constrained fusions yield shorter execution times compared to \textsc{Union} by restricting complex fusions that complicate subsequent FL rounds. GES maintains manageable execution times due to its one-shot strategy; the complexity would lie in initiating a new FL iteration from the intricate result generated by GES. Therefore, given that FedGES performs subsequent iterations, it is advantageous to limit the maximum number of edges it can add per iteration and opt for more restrictive fusions.

%
%
\section{Conclusions} \label{sec:conclusions}
We introduce Federated GES (FedGES), a novel Federated Learning approach for Bayesian Network (BN) structure learning in decentralized settings. FedGES utilizes the Greedy Equivalence Search (GES) algorithm to iteratively fuse the limited locally generated BN structures from individual clients, creating a unified global structure. The proposed controlled structural fusion in FedGES enhances consensus among different BNs learned by clients, offering a valuable solution in scenarios with larger BNs or numerous clients. This approach ensures data privacy by exchanging only network structures, thus avoiding the exposure of sensitive parameters. Our practical demonstrations over 14 BNs highlight FedGES's real-world applicability, consistently outperforming various federated and non-federated algorithms while maintaining the privacy of local data.

In future research, we plan to investigate the impact of client heterogeneity on the performance of FedGES, and its ability to handle non-IID data distributions among clients would also be valuable. Different strategies for non-uniform fusion in the server will also be investigated. A second line of research will address the process of parameter learning for the network in a federated manner, adding a second phase to FedGES. On the other hand, we could explore additional security mechanisms, e.g. considering the application of obfuscation processes such as differential privacy to secure the exchange of DAGs further. Studying in-depth the response of FedGES to different types of malicious attacks is also a promising line of research.


\subsubsection{Acknowledgements} The following projects have funded this work: TED2021-131291B-I00 and FPU21/01074 (MICIU/AEI/10.13039/501100011033 and ERDF, EU); PID2022-139293NB-C32 (MICIU/AEI/10.13039/501100011033 and European Union NextGenerationEU/PRTR); SBPLY/21/180225/000062 (Government of Castilla-La Mancha and ERDF, EU); 2022-GRIN-34437 (Universidad de Castilla-La Mancha and ERDF, EU).

This preprint has not undergone peer review or any post-submission improvements or corrections. The Version of Record of this contribution is published in Lecture Notes in Computer Science, vol 15244, and is available online at \href{https://doi.org/10.1007/978-3-031-78980-9\_6}{https://doi.org/10.1007/978-3-031-78980-9\_6}.

%
%
%
\bibliographystyle{splncs04}
\bibliography{biblio}

\begin{thebibliography}{10}
\providecommand{\url}[1]{\texttt{#1}}
\providecommand{\urlprefix}{URL }
\providecommand{\doi}[1]{https://doi.org/#1}

\bibitem{juanin2013GES}
Alonso, J.I., de~la Ossa, L., Gámez, J.A., Puerta, J.M.: Scaling up the Greedy Equivalence Search algorithm by constraining the search space of equivalence classes. International Journal of Approximate Reasoning  \textbf{54}(4),  429--451 (Jun 2013)

\bibitem{BarredoArrieta2020XAI}
Barredo~Arrieta, A., Díaz-Rodríguez, N., Del~Ser, J., et~al.: Explainable Artificial Intelligence (XAI): Concepts, taxonomies, opportunities and challenges toward responsible AI. Information Fusion  \textbf{58},  82–115 (Jun 2020)

\bibitem{Borunda2016BNSurvey}
Borunda, M., Jaramillo, O., Reyes, A., Ibarg\"{u}engoytia, P.H.: Bayesian networks in renewable energy systems: A bibliographical survey. Renewable and Sustainable Energy Reviews  \textbf{62},  32–45 (Sep 2016)

\bibitem{deCamposLearningBN2011}
de~Campos, C.P., Ji, Q.: Efficient Structure Learning of Bayesian Networks Using Constraints. Journal of Machine Learning Research  \textbf{12},  663–689 (Jul 2011)

\bibitem{chickering2002GES}
Chickering, D.M.: Optimal {Structure} {Identification} {With} {Greedy} {Search}. Journal of Machine Learning Research  \textbf{3},  507--554 (Jan 2002)

\bibitem{learningBNNP2004}
Chickering, D.M., Heckerman, D., Meek, C.: Large-Sample Learning of Bayesian Networks is NP-Hard. Journal of Machine Learning Research  \textbf{5},  1287–1330 (2004)

\bibitem{Drury2017BNSurvey}
Drury, B., Valverde-Rebaza, J., Moura, M.F., de~Andrade~Lopes, A.: A survey of the applications of Bayesian networks in agriculture. Engineering Applications of Artificial Intelligence  \textbf{65},  29–42 (Oct 2017)

\bibitem{gamezLearningBN2011}
Gámez, J.A., Mateo, J.L., Puerta, J.M.: Learning {Bayesian} networks by hill climbing: efficient methods based on progressive restriction of the neighborhood. Data Mining and Knowledge Discovery  \textbf{22}(1),  106--148 (Jan 2011)

\bibitem{heckerman1995BDeu}
Heckerman, D., Geiger, D., Chickering, D.: Learning Bayesian Networks: The Combination of Knowledge and Statistical Data. Machine Learning  \textbf{20},  197--243 (Aug 1995)

\bibitem{Jensen2007BN}
Jensen, F.V., Nielsen, T.D.: Bayesian Networks and Decision Graphs. Springer New York (2007)

\bibitem{SHD2009}
de~Jongh, M., Druzdzel, M.J.: A comparison of structural distance measures for causal Bayesian network models. In: Klopotek, M., Przepiorkowski, A., Wierzchon, S.T., Trojanowski, K. (eds.) Recent Advances in Intelligent Information Systems, Challenging Problems of Science, Computer Science series, pp. 443 -- 456. Academic Publishing House EXIT (2009)

\bibitem{Kim2019SHDMoral}
Kim, G.H., Kim, S.H.: Marginal information for structure learning. Statistics and Computing  \textbf{30}(2),  331--349 (Jul 2019)

\bibitem{Kyrimi2021BNSurvey}
Kyrimi, E., McLachlan, S., Dube, K., Neves, M.R., Fahmi, A., Fenton, N.: A comprehensive scoping review of Bayesian networks in healthcare: Past, present and future. Artificial Intelligence in Medicine  \textbf{117},  102108 (Jul 2021)

\bibitem{Laborda2023cGES}
Laborda, J.D., Torrijos, P., Puerta, J.M., G{\'a}mez, J.A.: A Ring-Based Distributed Algorithm for Learning High-Dimensional Bayesian Networks. In: Bouraoui, Z., Vesic, S. (eds.) Symbolic and Quantitative Approaches to Reasoning with Uncertainty. pp. 123--135. Springer Nature Switzerland, Cham (2024)

\bibitem{PGES2024}
Laborda, J.D., Torrijos, P., Puerta, J.M., Gámez, J.A.: Parallel structural learning of Bayesian networks: Iterative divide and conquer algorithm based on structural fusion. Knowledge-Based Systems  \textbf{296},  111840 (2024)

\bibitem{Leroy2019FLSurvey}
Leroy, D., Coucke, A., Lavril, T., Gisselbrecht, T., Dureau, J.: Federated Learning for Keyword Spotting. In: ICASSP 2019 - 2019 IEEE International Conference on Acoustics, Speech and Signal Processing (ICASSP). IEEE (May 2019)

\bibitem{Li_review_FL_applications_2020}
Li, L., Fan, Y., Tse, M., Lin, K.Y.: A review of applications in federated learning. Computers \& Industrial Engineering  \textbf{149},  106854 (2020)

\bibitem{mcmahan17aFL}
McMahan, B., Moore, E., Ramage, D., Hampson, S., Arcas, B.A.y.: {Communication-Efficient Learning of Deep Networks from Decentralized Data}. In: Singh, A., Zhu, J. (eds.) Proceedings of the 20th International Conference on Artificial Intelligence and Statistics. Proceedings of Machine Learning Research, vol.~54, pp. 1273--1282. PMLR (20--22 Apr 2017)

\bibitem{mian22aRFCD}
Mian, O., Kaltenpoth, D., Kamp, M.: Regret-based Federated Causal Discovery. In: Le, T.D., Liu, L., Kıcıman, E., Triantafyllou, S., Liu, H. (eds.) Proceedings of The KDD'22 Workshop on Causal Discovery. Proceedings of Machine Learning Research, vol.~185, pp. 61--69. PMLR (15 Aug 2022)

\bibitem{mian23aPERI}
Mian, O., Kaltenpoth, D., Kamp, M., Vreeken, J.: Nothing but Regrets — Privacy-Preserving Federated Causal Discovery. In: Ruiz, F., Dy, J., van~de Meent, J.W. (eds.) Proceedings of The 26th International Conference on Artificial Intelligence and Statistics. Proceedings of Machine Learning Research, vol.~206, pp. 8263--8278. PMLR (25--27 Apr 2023)

\bibitem{ng22aNOTEARS}
Ng, I., Zhang, K.: Towards Federated Bayesian Network Structure Learning with Continuous Optimization. In: Camps-Valls, G., Ruiz, F.J.R., Valera, I. (eds.) Proceedings of The 25th International Conference on Artificial Intelligence and Statistics. Proceedings of Machine Learning Research, vol.~151, pp. 8095--8111. PMLR (28--30 Mar 2022)

\bibitem{Nguyen2021FLSurvey}
Nguyen, D.C., Ding, M., Pathirana, P.N., Seneviratne, A., Li, J., Vincent~Poor, H.: Federated Learning for Internet of Things: A Comprehensive Survey. IEEE Communications Surveys \& Tutorials  \textbf{23}(3),  1622–1658 (2021)

\bibitem{penaConsensus2011}
Pe\~na, J.: Finding {Consensus} {Bayesian} {Network} {Structures}. The Journal of Artificial Intelligence Research (JAIR)  \textbf{42} (Jan 2011)

\bibitem{Puerta2021Fusion}
Puerta, J.M., Aledo, J.A., G{\'{a}}mez, J.A., Laborda, J.D.: Efficient and accurate structural fusion of Bayesian networks. Information Fusion  \textbf{66},  155--169 (Feb 2021)

\bibitem{Rieke2020FLSurvey}
Rieke, N., Hancox, J., Li, W., et~al.: The future of digital health with federated learning. npj Digital Medicine  \textbf{3}(1) (Sep 2020)

\bibitem{Saputra2019FLSurvey}
Saputra, Y.M., Hoang, D.T., Nguyen, D.N., et~al.: Energy Demand Prediction with Federated Learning for Electric Vehicle Networks. In: 2019 IEEE Global Communications Conference (GLOBECOM). IEEE (Dec 2019)

\bibitem{Scanagatta2019BNLearning}
Scanagatta, M., Salmerón, A., Stella, F.: A survey on Bayesian network structure learning from data. Progress in Artificial Intelligence  \textbf{8}(4),  425–439 (May 2019)

\bibitem{bnlearn}
Scutari, M.: Learning Bayesian Networks with the bnlearn R Package. Journal of Statistical Software  \textbf{35}(3),  1–22 (2010)

\bibitem{Silva2019FLSurvey}
Silva, S., Gutman, B.A., Romero, E., et~al.: Federated Learning in Distributed Medical Databases: Meta-Analysis of Large-Scale Subcortical Brain Data. In: 2019 IEEE 16th International Symposium on Biomedical Imaging (ISBI 2019). IEEE (Apr 2019)

\bibitem{Spirtes2001PC}
Spirtes, P., Glymour, C., Scheines, R.: Causation, Prediction, and Search. The MIT Press (2001)

\bibitem{Wang2023FedC2SL}
Wang, Z., Ma, P., Wang, S.: Towards Practical Federated Causal Structure Learning, p. 351–367. Springer Nature Switzerland (2023)

\bibitem{Yang2019FLSurvey}
Yang, W., Zhang, Y., Ye, K., et~al.: FFD: A Federated Learning Based Method for Credit Card Fraud Detection, p. 18–32. Springer International Publishing (2019)

\bibitem{Yao2021Causal}
Yao, L., Chu, Z., Li, S., Li, Y., Gao, J., Zhang, A.: A Survey on Causal Inference. ACM Transactions on Knowledge Discovery from Data  \textbf{15}(5),  1–46 (May 2021)

\bibitem{Zhang_survey_fl_KBS_2021}
Zhang, C., Xie, Y., Bai, H., Yu, B., Li, W., Gao, Y.: A survey on federated learning. Knowledge-Based Systems  \textbf{216},  106775 (2021)

\end{thebibliography}

\end{document}